# Effects of data ambiguity and cognitive biases on the interpretability of machine learning models in humanitarian decision making

**David Paulus**[1], **Gerdien de Vries**[2] and **Bartel Van de Walle**[3]
[1]HumTechLab, Delft University of Technology
email: d.paulus@tudelft.nl
[2]Organization and Governance Section, Delft University of Technology
email: G.deVries-2@tudelft.nl
[3]HumTechLab, Delft University of Technology
email: B.A.vandeWalle@tudelft.nl

**Abstract**

The effectiveness of machine learning algorithms depends on the quality and amount of data and the operationalization and interpretation by the human analyst. In humanitarian response, data is often lacking or overburdening, thus ambiguous, and the time-scarce, volatile, insecure environments of humanitarian activities are likely to inflict cognitive biases. This paper proposes to research the effects of data ambiguity and cognitive biases on the interpretability of machine learning algorithms in humanitarian decision making.

## 1 Introduction

Humanitarian response comprises a wide range of activities conducted by a multitude of actors in diverse contexts [1]. Activities include the search and rescue of wounded and deceased, delivery of aid to the affected, camp management and inter-organizational coordination [23]. Actors are the affected communities themselves, local and national groups and organizations, governmental agencies, the private sector, the military, international non-governmental organizations and the United Nations, as well as digital volunteers [2]. Humanitarian contexts can be categorized through locus, type and extent of disasters and crises as well as by the social, cultural and political environments they occur in [12].

Implementing organizations of humanitarian activities mainly operate through funds from donor country governments [5]. From that perspective, allocation decisions are being made on three levels: on a donor level regarding if and what amounts of funds are to be allocated to what humanitarian context and recipient organization. On an organizational headquarter level regarding what resources are to be allocated to what field operations. And on a field level regarding what resources are to be allocated to local partners and how to allocate aid to groups of affected people.

On all three levels, decision makers are challenged by data ambiguities and are influenced by cognitive biases but have to make decisions anyway. Often, decisions are made in the absence of reliable data and - for example due to stress, time-, and resource-constraints - under the influence of cognitive biases [3].

**An example case**

Yemen experiences the worst humanitarian crisis of today [25]. Activities for emergency food assistance have received the largest funds from international donor countries compared to activities in other sectors (e.g. health, education, protection) in Yemen (Figure 1).

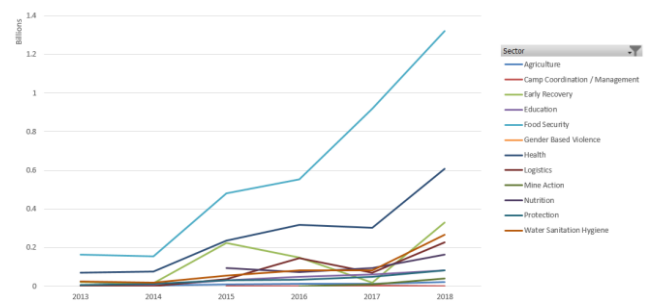

Figure 1. Funds donated to humanitarian sectors in Yemen. In USD. Source: UN OCHA Financial Tracking Service.

But funds have not been sufficient. And in cases where humanitarian organizations cannot reach all affected people, they have to prioritize certain groups and areas over others [28].

To support allocation decisions, machine learning can support the analysis of population data in combination with food security indicators and data on the changing conflict in Yemen. Ideally, algorithms predict how many rations will be needed when and where, thereby increasing the number of people reached, mitigating human suffering and reducing operational costs [17].

The efficiency of a machine learning algorithm depends on the data it is trained with [20]. If the data is not accurately representing reality or if it is lacking important attributes, these shortcomings are cascading into the model and its predictions [22]. In humanitarian response, data is often problematic [19]. But even if the data is reliable, biases can lead decision makers to ignore or misinterpret the re-sults of the algorithm, or even collect and input data in a biased way. The time-scarce, stressful situation in Yemen is likely to inflict cognitive biases that can affect decision makers in e.g. deciding over whether to conduct a needs assessment, what data to trust and how to act on a prediction of a model.

## 2 Previous work

Data ambiguity has been defined as one aspect of data quality that leads to false interpretations because of inaccuracies within the data [6]. Data ambiguity is thus one of a number of aspects that make up data quality. [26] proposed a conceptual framework of data quality in which they categorized data characteristics into four categories: intrinsic, contextual, representational and accessibility data quality. Ambiguity, in general terms, is the quality of being open to more than one interpretation, and closely related to inexactness. This paper uses the term data ambiguity to summarize the data quality characteristics of accuracy, relevancy, timeliness, completeness and puts a particular focus on interpretability.

Interpretability of data for humanitarian organizations was studied by [4], in their study on decision makers' information needs in the response to Typhoon Haiyan. The authors argued that the heterogeneity of data caused confusion between different organizational levels and suggested more standardization of data and a comprehensive ontology that enables a "shared conceptualization" of the disaster situation. They further stressed the role of coordination to provide "relevant, accurate and timely" data to all actors in the response. They found "necessary but competing agendas" between organizations' headquarters and field offices. Headquarters required frequent, accurate reports which field staff could not deliver and which impeded their own planning and activities on the ground. Their findings show a strong effect of data ambiguity on intra-organizational understanding.

The lack of reliable data and the uncertainty it creates, leaves room for heuristics and cognitive biases humanitarian actors fall back to in their decision making processes. The idea of cognitive biases and that they influence individuals through heuristics so that their judgments differ from purely rational thinking, was introduced by [10]. Biases are often characterized as a byproduct of information processing limitations: because of a lack of time or capacities, people use these mental shortcuts to judge and decide. There is "broad consensus that human decision making relies on a repertoire of simple, fast, heuristic decision rules to be used in specific situation" [8, p. 729]. While most studies on cognitive biases label them as fallacies, [7] showed that simple decision making approaches can perform equally or better compared to sophisticated approaches that try to gather and process all available information. And according to [8], many biases may have developed to favor "inexpensive, frequent errors rather than occasional disastrous ones" [8, p. 731].

While cognitive biases in humanitarian emergencies have only been hypothesized yet, evidence exists for cognitive biases in other high-risk situations. Examples are studies on group behavior and decision making in physically and mentally extremely challenging events, for example [21, 15]. From those studies, some biases appear more dominant than others and include sunk cost fallacy, overconfidence, recency bias and confirmation bias. These are also part of the list of biases postulated by [3] that might be influencing decision makers in humanitarian emergencies. [3] argues these biases might be influential because of the "stress and pressure, distorted, lacking and uncertain information" in humanitarian emergencies. And that decision makers in the humanitarian sector develop coping strategies to deal with the high number of decisions they have to conduct in short periods of time.

To the best of our knowledge, effects of data ambiguity and cognitive biases on the interpretability of machine learning models in humanitarian response have not been studied. So far, machine learning has been applied and studied in a number of applications within the humanitarian sector. [19] found that high accuracy levels can be achieved by applying machine learning models on mapping tasks for refugee settlements. And [16] proposed a hybrid human-machine learning approach to analyze large amounts of satellite imagery data in disaster contexts. And a number of scholars utilized machine learning approaches to assess social media data during emergencies [9]. [22] discussed cognitive biases and their potential effects on the interpretations by human analysts of rule-based machine learning models. And [24] found that machine learning algorithms can perform better in classifying small datasets when biases are artificially coded into the algorithms.

We therefore propose a set of research questions to investigate the effects of data ambiguity and cognitive biases on the interpretability of machine learning models in humanitarian decision making in the next section.

## 3 Proposed Research

Looking back at the three-level perspective on the humanitarian sector, and taken into account the evidence of conflicting understandings between these levels, the proposed research should investigate the following questions for each level individually.

*What is a suitable measure for interpretability [18] of machine learning models in humanitarian decision making?*
Machine learning analysts and decision makers in donor country agencies likely follow different rules to achieve different results than their counterparts in recipient organizations' headquarters and field offices. Also their professional backgrounds and trainings might vary, leading to different approaches to operationalize machine learning models and assess and value their results.

*What characteristics of data ambiguity influence interpretability of machine learning models in humanitarian decision making?*
Data characteristics – including accuracy, timeliness, completeness, trustworthiness, format – might be valued differently throughout the three levels and are highly context-dependent. During search and rescue operations, decision makers might favor fast results over completeness and accuracy, to save time and accelerate relief operations.

*What cognitive biases influence interpretability of machine learning models in humanitarian decision making?*
Above mentioned literature points to potential biases that analysts and decision makers might be prone to adopt, which can have positive and negative effects on interpretability. Studies within cognitive psychology, however, have found many more biases that are worthwhile to investigate in humanitarian decision making.

*What interventions support positive effects and mitigate negative effects on the interpretability of machine learning models in humanitarian decision making?*
Effects that are discovered through the previous questions, might then be available to be strengthened or weakened through individual or organizational interventions. Sensemaking [27] was previously suggested as an organizational measure to adapt to uncertain and ambiguous humanitarian environments [14]. Debiasing, which often entails a training component on personal awareness, also holds potential to interfere with either positive or negative effects on interpretability [13, 11].